# HYPONYMY EXTRACTION OF DOMAIN ONTOLOGY CONCEPT BASED ON CCRFS AND HIERARCHY CLUSTERING*

Qiang Zhan[1,2] and Chunhong Wang[2]

[1]School of Computer Science and Technology, Beijing Institute of Technology, Beijing, China

[2]Dept. of Computer Science, Yuncheng University, Shanxi, China

### ABSTRACT

*Concept hierarchy is the backbone of ontology, and the concept hierarchy acquisition has been a hot topic in the field of ontology learning. this paper proposes a hyponymy extraction method of domain ontology concept based on cascaded conditional random field(CCRFs) and hierarchy clustering. It takes free text as extracting object, adopts CCRFs identifying the domain concepts. First the low layer of CCRFs is used to identify simple domain concept, then the results are sent to the high layer, in which the nesting concepts are recognized. Next we adopt hierarchy clustering to identify the hyponymy relation between domain ontology concepts. The experimental results demonstrate the proposed method is efficient.*

## 1.INTRODUCTION

Automatic acquisition of hyponymy relations is a basic problem in knowledge acquisition from text, and commonly used in the construction and verification of ontology, knowledge base and lexicon. Give concepts C1 and C2, if the extension of C2 includes the extension of Cl, then C1 and C2 can be thought to have the hyponymy relation, and we can say C2 is the hyponymy concept of C1, and C1 is the hyponymy concept of C2. Lower and an upper concepts Cl、C2 form a hyponymy relation and we denote the hyponymy relation as ISA(C1,C2) . A simple way to judge if a ISA(C1,C2) relation is correct or not is to judge if the sentence "C1 is a kind of/ is one of  C2" is correct. For example, "compact disk is a kind of computer hardware" is correct ,that is, ISA(compact disk , computer hardware) is correct.

"is-a" relationship of domain terminology is a basic and important problem in the construction of domain ontology. So Automatic acquisition of "is-a" relations can be used to construct the classification of domain ontology. This paper propose a method base on CCRFs to automatic acquire the "is-a" relation between domain concepts. First we carry on semantic analysis, then utilize the result of semantic analysis to aggregate the domain concepts, at last, acquire "is-a" relationship through calculating the co-occurrence.

# 2.RELATED WORKS

For the acquisition of "is-a" relationship between domain concepts, commonly used methods are as follows: template method, lexicon driven method, clustering method and hybrid method[1,2,3,4].

Lexicon method is to acquire the relationship between the concepts according to synonym, antonym defined in the lexicon. Nakaya and his colleague[5] use WordNet to acquire the classification relationship between the concepts. The basic thinking is as follows: first confirm the category of the relationship, then take the concept semantics as center, after that extract the specific relationship according to the search depth designated by customer. This method currently can extract the synonymous, antonymous and "is-a" relationships.

Through analysis of domain relevant documents, the template method summary some linguistic mode with high frequent occurrence, and take these modes as the rules, then judge whether the word sequence in the document match a certain mode. If match, then a corresponding relation can be recognized[6]. These modes can either be defined manually, or obtained through machine learning from corpus. Hearts[4] Earlier proposed a hierarchy acquisition method base on template. The purpose of this proposed method is to solve two problem: the preprocess of the knowledge and applied need for a large scale of documents. Hearst first define a set of syntax template which is easy to be recognized, the template has a high occurrence frequency, cover most of the domains, and it can be able to recognize the relationship, after that he design a method which can be able to automatically recognize the template, which make this method apply to different domains possible.

Caraballo[7] and his colleague utilize conjunction to acquire noun, construct the noun characteristic vector using the conjunction relationship and apposition relationship of noun phrase in the context, utilize the aggregation to obtain the hyponym relationship between the nouns. Except for that, Caraballo also improve the method of Hearst, that is, label the inner node in the hierarchy structure. Weaknesses also exist in the method based on template: Low precision. Whether the template is complete or not affect the result of the extraction severally. The extracted relationship is on the level of words, not on the level of concepts, so there exist a problem of word sense disambiguation.

Fisher[8] proposed a method based on vector clustering-COBWEB. The input object is duplex{attribute, attribute value}. COBWEB use the taxonomy tree to construct hierarchy clustering. Each point of taxonomy tree correspond to a concept, containing a description about the concept.

Cimiano[9] proposed a hybrid method for the hyponym relation extraction. They combined several characteristics of WordNet、the template of Hearst, and use machine learning method to construct classifier. They use different classifier to recognize the hyponym relation.

The method based on the clustering analysis utilize the syntax distance between the concepts to cluster the concepts. The concepts in the same cluster have the near semantic relationship. The result of cluster is the taxonomy of the concepts. This paper describes a hyponymy extraction algorithm of domain ontology concepts based on CCRFs and hierarchy clustering. In section 3, we describe the motivation. In section 4, we summary the process and related concepts of our

This work is supported by National Nature Science Foundation Of China under Grant 61371194

algorithm. Section 5 describes experiment, including corpus selection and evaluation method, result analysis. In section 6, we conclude our work and future research.

# 3.MOTIVATION

Domain ontology learning need to utilize the hyponym relationship to construct ontology hierarchy relationship. This paper aims to propose a method to automatically acquire the relationship between the domain concepts. The method has the characteristic as following: Realize automatic extraction of domain concept hyponym relationship. The purpose of this paper is to construct the domain ontology, the method proposed in this paper should be able to extract the hyponym relationship from domain concepts.

# 4.PROPOSED APPROACH

In this section we present our proposed extraction system. According to the different purpose, we can divide the process into two parts:

- Domain concept reorganization based on Cascade conditional random fields

  CRFs has long-distance dependencies, and CRFs avoid the label bias problem, a weakness exhibited by maximum entropy. In addition, it can get precise global optimization solution. CCRFs is the improvement of CRFs, and CCRFs can recognize the nesting concepts better. So we use CCRFs to construct the model to recognize the concept on the training corpus, first we use CRFs to recognize the simple concept, then we send the result of the first layer to the high layer, finally we use the high layer to recognize the complex nesting concepts.

- Hyponymy relation extraction based on hierarchy clustering

  The context information of one word contains abundant the semantic information about this word, one word often has the different semantics in the different context, so we can use the words in the context environment of the target word to construct a vector to represent the target word.

## 4.1.Domain concept reorganization based on Cascade conditional random fields
### 4.1.1.Conditional random fields

Conditional Random Fields (CRFs), a statistical sequence labeling model, was first introduced by Lafferty et al. (2001) [9]. CRFs are also undirected graphical models, used to calculate the conditional probability of values on designated output nodes given values assigned to other designated input nodes. Comparing with maximum entropy model, CRF models overcome the label bias problem and can trade off decisions at different sequence positions to obtain a globally optimal labeling. Let $O = (o_1, o_2, ..., o_n)$ denote an observation value sequence, such as the input Chinese word sequence. Let $S = (s_1, s_2, ..., s_n)$ denote an output state sequence, such as an output tagging sequence. The conditional probability of a label sequence $S$ given an input sequence defined by CRF is:

$$p(s|o) = \frac{1}{Z(o)} \exp(\sum_i \sum_k \lambda_k f_k(s_{i-1}, s_i, o, i)) \qquad (1)$$

Where $Z(o)$ is a normalization factor over sequences.

$$Z(o) = \sum_{s \in S} \exp(\sum_i \sum_k \lambda_k f_k(s_{i-1}, s_i, o, i)) \qquad (2)$$

$f_k(s_{i-1}, s_i, o, i)$ is an arbitrary feature function over its arguments. $\lambda_k$ is a learned weight for each feature function, can be obtained through training.

### 4.1.2. Cascade conditional random fields model

Due to the significant characteristic of computer domain, it is very common to find some complex terms in the corpus, Such as "备份域控制器"(Backup Domain Controller). Complex terms are more suitable to be recognized after the recognition of simple terms. So, it is necessary to introduce cascaded conditional random fields into recognizing compound terms.

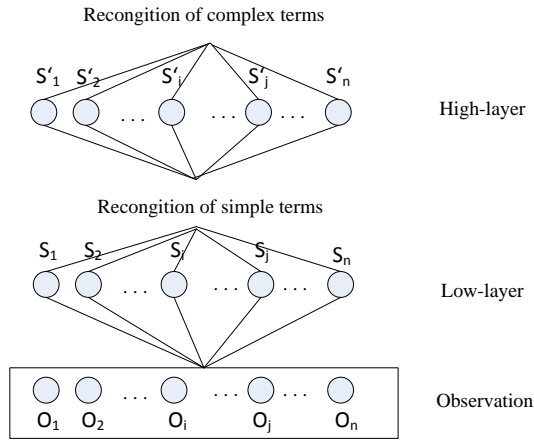

Fig.1. the model of cascaded conditional random fields

The model of cascade conditional random fields is shown as Fig 1. The system includes two phases: a low-layer CRFs phase which is used to extract simple domain terms, a high-layer CRFs phase which is used to extract complex domain terms.

### 4.1.3. Cascade conditional random fields algorithm

- Ccorpus pre-processing

We crawl a certain scale of webpage from WWW, remove the HTML label and get the plain text. This paper changes the task of term recognition into the task of sequence labeling. We adopt the popular BIO method to represent the result of sequence labeling, that is, B represents the beginning of a term; I represents the rest part of the term; O represents the other word which is not a term.

In our algorithm, corpus preprocessing appears two times: the preprocessing in the low-layer of CRFs and the preprocessing in the high-layer of CRFs. This first preprocessing pay close attention to simple terms, such as "如何安装备份域控制器"(how to install a Backup Domain Controller). After labeling, it is "如何/O 安装/O 备份/B 域/I 控制/B 器/I". After the process of the low-layer, we get a sequence labeling of Chinese words. In the second corpus preprocessing, we take complex terms as objective. For example, "如何安装备份域控制器" (how to install a Backup Domain Controller), after labeling, it is "如何/O 安装/O 备份域/B 控制器/I"

- Feature selection

In this paper, feature is defined as a kind of regulation to describe the regularity of data and the statistics characteristic. They can be defined without limitation, but more is not better on feature. The features must be representative and effective. For CRFs model, it is very important to select appropriate features to express the complex natural language phenomenon. Considering the feature of the field of computer domain, we adopted five features as follows:

Word (Word): As a component of a term, words themselves provide some domain knowledge.
Part-of-speech (POS): In the field of computer, term has a certain POS composing rules. Most of them are noun phrase. So POS is an important characteristic for term identification.

The length of word (WordLen): It is the number of characters in a word. According to the length of word, we can judge if the current word is a part of a term.

Left information entropy (Lien)

Right information entropy (Rien): Left and right information entropies are important measures for judging if a word is the borderline of a term, so it can be said to be a term's unithood.

- Feature templates

After selecting the features, then feature templates should be designed. According to the feature templates, CRFs model generate a series of characteristic function to stipulate how the characteristics are used. Feature templates are the comprehensive consideration of the special information and the information location. For the first layer, the feature templates for the above five features are shown as table 1.

Table 1. feature templates in the field of computer

| Type | Feature templates |
|------|-------------------|
| Word | W(-2)，W(-1)，W(0)，W(+1)，W(+2)，W(-2)+W(-1)，W(-1)+W(0)，W(0)+W(+1)，W(+1)+W(+2) |
| POS | P(-2)，P(-1)，P(0)，P(+1)，P(+2)，P(-2)+P(-1)，P(-1)+P(0)，P(0)+P(+1)，P(+1)+P(+2) |
| WordLen | WL(-2)，WL(-1)，WL(0)，WL(+1)，WL(+2),WL(-2)+WL(-1)，WL(-1)+WL(0)，WL(0)+WL(+1)，WL(+1)+WL(+2) |

| Lien | L(0) |
|------|------|
| Rien | R(0) |
| Hybrid feature | W(-2)+P(-2)，W(-1)+P(-1)，W(0)+P(0)，W(+1)+P(+1)，W(+2)+P(+2) |

In the Table 1, W represents the word itself, P represents the POS, WL represents the length of word, L represents the left information entropy, and R represents the right information entropy. "+" represents the combination of features. The window of the feature templates is five, that is, the features of the next two words and the previous two words will be considered to add the context information of the current word. W(0) represents the feature of the current word, W(-1) and W(-2) represent the features of the previous two words. W(+1) and W(+2) represent the features of the next two words. W(-1)+P(-1) represents the affection of the previous word and its POS to the current word.

## 4.2.Hyponymy relation extraction based on hierarchy clustering
### 4.2.1.Vector representation of domain concept

The first problem that the hierarchy clustering need to solve is how to use a vector to represent a concept word. WordSpace model is a multi-dimension space model which is constructed with vector . In it, each vector represent a concept word. One word can be represented by counting the word co-occurrence in the corpus, these co-occurrence can construct a vector. One word context information contain abundant the semantic information about the word. One word always has the different semantics in the different context, so we can use the words in the context environment of the target word to construct a vector to represent the target word. Each element in the vector has two component : the context words of the target word and the frequency of the word. This vector is defined as the following:

$$\vec{c} = ((w_1, f_1), (w_2, f_2), \ldots, (w_n, f_n)) \qquad (3)$$

In the equation, $c$ is the target word, $w_i$ is the co-occurrence word of $c$, $f_i$ is the co-occurrence frequency of $w_i$. Vector $\vec{c}$ is defined as the word space vector of the target word. The context words generally are selected in a fixed range of the target word. This range is called window, and be represented as [a,b], that is, the left a words of the target word and the right b words of the target word, the size of the window is adjusted according to the concrete circumstance.

### 4.2.2.Similarity matrix

The step of constructing the word space is as following:

- After the extraction of domain concepts,   construct the domain concept collection. The size of the collection is N.
- For every word W in the collection, construct the word vector. Count the frequency of the all context co-occurrence in the corpus, we get

$$\vec{w} = ((w_1, f_1), (w_2, f_2), \ldots, (w_n, f_n)) \qquad (4)$$

We choose the N words with the highest occurrence to represent W, we get the W vector $\vec{w}$. After every word in the concept collection is represented with a vector, we get the word space:

$$\vec{c} = \{\vec{w}_q | q = 1, \dots, n\} = \{\vec{w}_1, \vec{w}_2, \dots, \vec{w}_n\} \quad (5)$$

- Calculate the similarity between the all vector respectively, we can get the semantic similarity matrix between the concepts. Each value in the similarity matrix represents a similarity between a couple of the concepts.

$$w_{ij} = \{sim(\vec{w}_i, \vec{w}_j) | i < j \text{且 } i, j = 1, \dots, n\} \quad (6)$$

Similarity matrix is shown as following：

$$\begin{bmatrix} 0 & sim(\vec{W}_1, \vec{W}_2) & sim(\vec{W}_1, \vec{W}_3) & \cdots & sim(\vec{W}_1, \vec{W}_n) \\ & 0 & sim(\vec{W}_2, \vec{W}_3) & \cdots & sim(\vec{W}_2, \vec{W}_n) \\ & & 0 & \cdots & sim(\vec{W}_3, \vec{W}_n) \\ & & & \ddots & \vdots \\ & & & & 0 \end{bmatrix} \quad (7)$$

### 4.2.3. Similarity measure

In the process of hierarchy clustering, we need to calculate the distance between the different classes to finish the aggregation of clustering. The distance between the classes is realized through calculating the distance between the elements in the class. The element here means the concept, So we need to calculate the distance between the concepts. Because the concept adopt the vector notion, so the task of calculating the similarity between the concepts can be changed to the task of calculating the similarity between the vectors. Thereby the vector similarity calculation is relatively simple. For two random concepts, they can be represented by vector as $\vec{w}_1$、 $\vec{w}_2$.

$$\vec{w}_1 = X_1/a_1, X_2/a_2, X_3/a_3, X_4/a_4 \quad (8)$$

$$\vec{w}_2 = X_2/b_2, X_3/b_3, X_4/b_4, X_5/a_5 \quad (9)$$

In the equation, $(X_1, \dots, X_5)$ represents the context word, $(a_1, \dots, a_4)$, $(b_2, \dots, b_5)$ represents the frequency of the context words.

Due to the different words in the $W_1$ and $W_2$, So if we want to calculate the similarity between $W_1$ and $W_2$, first expand the word to get them have the same words. Add the word which doesn't occur in the $W_1$ to $W_1$ and set the frequency 0.

$$\vec{w}_1' = X_1/a_1, X_2/a_2, X_3/a_3, X_4/a_4, X_5/0 \quad (13)$$

$$\vec{w}_2' = X_1/0, X_2/b_2, X_3/b_3, X_4/b_4, X_5/a_5 \quad (14)$$

We use cosine method to calculate the similarity between two vectors, and get:

$$sim(\vec{w}_1, \vec{w}_2) = \cos(\vec{w}_1, \vec{w}_2) = \frac{\sum_{a \in \vec{w}_1, b \in \vec{w}_2} ab}{\sqrt{\sum_{a \in \vec{w}_1} a^2 \sum_{b \in \vec{w}_2} b^2}} \quad (15)$$

**4.2.4.The algorithm of the concept hierarchy clustering**

Input: concept collection $C = \{c_i | i = 1, \dots, n\}$, in this equation, $c_i$ is concept, n is the size of the concept collection. The threshold of similarity is t.

Output: clustering collection $K = \{k_i | i = 1, \dots, m\}$, in the equation, m is the number of the clustering.

- Initiation, separate every concept into a single cluster.
- Calculate the similarity between clusters respectively
- Choose the biggest two clustering $k_i$, $k_j$, if the similarity is bigger than the input the threshold t, then get them into one class, then go to 2, if their similarity is smaller than t, then go to 4
- End

# 5.EXPERIMENT AND RESULT
## 5.1.The Corpus Collection

The corpus is collected from computer domain website. It includes 200 documents containing 107,821 characters. We choose 100 documents from domain corpus and take it as training corpus, and take the rest 100 documents as open testing corpus. We adopt precision (P), recall(R) and F-measure as the evaluation criteria. The P, R and F-measure are defined as follows:

$$P = \frac{\text{the number of correct hyponym relationship extracted}}{\text{the number of hyponym relationship recognized}} \quad (16)$$

$$R = \frac{\text{the number of correct hyponym extracted}}{\text{the number of hyponym relationship in the corpus}} \quad (17)$$

$$F - score = \frac{2 \times P \times R}{P + R} \quad (18)$$

## 5.2.Experimental results and analysis

In order to evaluate the proposed method objectively, we carry on two experiments. First we use the different window size and word vector dimensions to test the proposed method. The result is shown as Table 2. Next we compare our method with the method based on pattern clusters[10]. The result is shown as Table 3.

Table 2     the influence of different word vector parameter for the extraction of hyponymy relation

| | | | | | | | |
|---|---|---|---|---|---|---|---|
| parameter | the window size | $[2, 2]$ | $[4, 4]$ | $[8, 8]$ | $[8, 8]$ | $[12, 12]$ | $[12, 12]$ |
| | Word vector dimension | 4 | 8 | 8 | 12 | 8 | 12 |

| number of the relation extracted | 216 | 218 | 226 | 240 | 233 | 236 |
|---|---|---|---|---|---|---|
| correct relation number | 98 | 114 | 149 | 175 | 166 | 147 |
| number of manual statistic relation | 221 | 221 | 221 | 221 | 221 | 221 |
| recall (%) | 44.34 | 51.58 | 67.42 | 79.18 | 75.34 | 66.51 |
| precision (%) | 45.37 | 52.29 | 65.92 | 72.91 | 71.24 | 62.28 |

Table 3    Comparison of our method and the method based on pattern clusters

| method | P | R | F-score |
|---|---|---|---|
| our method | 72.91 | 79.18 | 75.91 |
| the method base on pattern clusters | 73.36 | 67.81 | 70.49 |

From the table 2, we can see the recall and precision is the best when the window size is [8,8], dimension is 12. If the window is small, it can not express the complete concept context information, but if the window is large, it contains too much futile information, affect the result of the extraction. The dimension is small, it can not express the complete context semantics of the context too, if the dimension is large, the noise is large, then affect the extraction. According to the actual situation, we can adjust the window size and dimension.

From the table 3, the identification result of our method is better than the method based on pattern clusters. Compared to the method based on rules, we can see the precision of the method based on the hierarchy clustering is low, and the recall is higher. Two main reason lead to the circumstance:

- The method based on the rules extract the hyponymy relation through the match of rules, but the rules in the rules repository are generally verified and have high verification. So once it match successfully, the fault of the relation extracted is small, that is, the precision is high.
- The recall of the method based on the rules is low because the rules in the rules repository is not complete, and it can not express the complete relation and rules. Some relation occurring in the corpus can not match the corresponding the rule in the repository, and can not be extracted. But the method based on the hierarchy clustering is only a method based on the statistics, and as long as the scale of the corpus is large enough, the complete relation can be extracted completely

# 6.CONCLUSION

Hyponymy relation extraction of domain concepts has been a hot research in the information processing field in recent years. This paper use CCRFs to extract the domain concepts, the low layer is to extract the domain simple concepts and the high layer is to extract the domain complex concepts. Then we use the hierarchy clustering to extract the hyponym relation between domain concepts. The experimental results show the proposed method is efficient. For future work, we are now looking at how to improve the precision of the hyponymy extraction based on our method.

## Authors:

Qiang Zhan (1975-), male, Doctor of Beijing institute of technology, my main interest is in the field of Natural language processing.

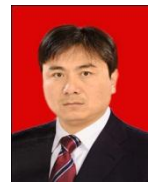

Chunhong Wang(1965-), female, professor, research field is Natural language processing.